\pdfoutput=1
\documentclass[11pt]{article}

\usepackage[final]{acl}

\usepackage{times}
\usepackage{latexsym}
\usepackage{booktabs}
\usepackage{subcaption}
\usepackage{tabularx}
\usepackage{tcolorbox}
\usepackage{multirow}
\usepackage{pifont}
\usepackage{enumitem}

\usepackage[T1]{fontenc}
\usepackage[utf8]{inputenc}

\usepackage{microtype}

\usepackage{inconsolata}

\usepackage{graphicx}

\definecolor{darkred}{rgb}{0.5, 0.0, 0.0}
\definecolor{darkgreen}{rgb}{0.0, 0.5, 0.0}
\definecolor{whitesmoke}{rgb}{0.96, 0.96, 0.96}

\newenvironment{prompt}{%
  \begin{tcolorbox}[
    arc=0pt,
    outer arc=0pt,
    boxrule=0.8pt,
    colback=whitesmoke,
    fontupper=\ttfamily,
    colframe=gray!50!black,
    sharp corners=southwest
  ]%
}{%
  \end{tcolorbox}%
}

\title{How well do LLMs reason over tabular data, really?}

\author{Cornelius Wolff \\
  Centrum Wiskunde \& Informatica \\
  Amsterdam \\
  \texttt{cornelius.wolff@cwi.nl} \And
  Madelon Hulsebos \\
  Centrum Wiskunde \& Informatica \\
  Amsterdam \\
  \texttt{madelon.hulsebos@cwi.nl}
}

\begin{document}
\maketitle

\begin{abstract}
Large Language Models (LLMs) excel in natural language tasks, but less is known about their reasoning capabilities over tabular data. Prior analyses devise evaluation strategies that poorly reflect an LLM's realistic performance on tabular queries. Moreover, we have a limited understanding of the robustness of LLMs towards realistic variations in tabular inputs. Therefore, we ask: \textit{Can general-purpose LLMs reason over tabular data, really?}, and focus on two questions 1) are tabular reasoning capabilities of general-purpose LLMs robust to real-world characteristics of tabular inputs, and 2) how can we realistically evaluate an LLM's performance on analytical tabular queries?\\
Building on a recent tabular reasoning benchmark, we first surface shortcomings of its multiple-choice prompt evaluation strategy, as well as commonly used free-form text metrics such as SacreBleu and BERT-score. We show that an LLM-as-a-judge procedure yields more reliable performance insights and unveil a significant deficit in tabular reasoning performance of LLMs. We then extend the tabular inputs reflecting three common characteristics in practice: 1) missing values, 2) duplicate entities, and 3) structural variations. Experiments show that the tabular reasoning capabilities of general-purpose LLMs suffer from these variations, stressing the importance of improving their robustness for realistic tabular inputs.\footnote{Code: \href{https://github.com/trl-lab/tabular-robustness}{github.com/trl-lab/tabular-robustness}}
\end{abstract}

\section{Introduction}

Large Language Models (LLMs) are intended for general-purpose usage and particularly excel on natural language tasks represented in text~\cite{liang2023helm}. In organizations, another common modality for data analysis and decision-making, is tabular data, for which recent studies have shown promising performance of LLMs as well~\cite{fang2024large}. Structural analysis to understand to what extend the reasoning capabilities of LLMs pertain, \textit{realistically}, in more complex tabular reasoning tasks, such as analytical aggregations, is still lacking. Without reliable knowledge of their failure modes on tabular inputs and tasks, though, we risk unwarranted usage of these models in practice and delayed development of more robust capabilities.

Surfacing the reasoning capabilities on tabular tasks is, however, not straightforward. Most studies adopt free-form text metrics, which hardly capture reliable reasoning accuracy due to different formatting of the ground-truth answers, particularly of analytical questions, versus long-form LLM-generated responses~\cite{ji2024target, xu2023critical}. The alternative of forcing certain output formats~\cite{sui2024table} yields a limited understanding of the open-form reasoning performance and is prone to parsing errors, while another alternative of including the ground-truth answer in a multiple-choice prompt~\cite{qiu2024tqa} leaks ground-truth answers into the prompt compromising its reliability. Beyond realistic evaluation procedures, in order to use LLMs in a reliable manner for tabular reasoning tasks, it is important to understand how well they can handle the characteristics of tabular data inputs~\cite{cong2023observatory, singha2023tabular} as encountered in practice.

To close these gaps, we first address the question: \textit{how can we realistically evaluate an LLM's performance on analytical tabular queries?} We examine the limitations of existing evaluation metrics, such as SacreBleu \cite{post-2018-call} and BERT-score \cite{Zhang2020BERTScore:}, as the distributions are these among correct and incorrect answers are inseparable. Instead, we propose using the LLM-as-a-judge evaluation method~\cite{zheng2023judging} for more reliable performance insights and show, through calibration with human annotations, that the LLM-as-a-judge provides a reliable signal of tabular reasoning performance. Using this evaluation procedure, we unveil a significant gap in tabular reasoning accuracy than previously found in the existing TQA-Bench benchmark for tabular reasoning~\cite{qiu2024tqa}.

Second, through comprehensive analysis we answer the question: \textit{are LLMs robust to real-world characteristics of multi-table inputs?} Building on the TQA-Bench, we first improve the validity of the task queries and downscale multi-table inputs to gain more fine-grained insights. We then inspect the robustness of LLMs against characteristics of tabular inputs as commonly found in practice. Specifically, we formalize the following three characteristics: missing values, duplicate entities, and structural variations. We show that most LLMs are not as robust to, and insufficiently acknowledge the presence of, such quality issues or anticipated variations in multi-table inputs, highlighting the need for more robust models for tabular reasoning.

\noindent We make the following concrete contributions:
\begin{itemize}[leftmargin=0.5cm, topsep=3pt]
    \itemsep0em
    \item We concretize the limitations of free-form text metrics and show the reliability of using an LLM-as-a-judge for evaluating open-ended responses for tabular reasoning tasks.
    \item We extend the TQA-Bench benchmark with tabular inputs reflecting typical real-world characteristics of tabular data: missing values, duplicate entities, and structural variations.
    \item We surface the shortcomings of LLMs to account for realistic variations in tabular data of varying sizes, providing more fine-grained insights into their scalability and robustness.
\end{itemize}

\section{Related Work}

\paragraph{Analysis of Tabular Reasoning Capabilities}
The TQA-Bench~\cite{qiu2024tqa} examines multi-table reasoning capabilities with LLMs over various query complexities. We complement the TQA-Bench by using it as a base for our evaluation, and integrating common properties in tabular data, such as missing values, to study the robustness of multi-table reasoning capabilities of LLMs. Similarly, \citet{sui2024table} considers structural understanding capabilities by evaluating the accuracy of LLMs in basic tasks such as row/column retrieval. The QATCH benchmark~\cite{papicchio2023qatch} evaluates tabular representation learning models specialized for SQL-centric tasks and mainly focuses on SQL-based evaluations. While the QATCH benchmark considers enterprise-centric evaluation tasks and inputs, it does not surface robustness for real-world properties.
Earlier work by \citet{cong2023observatory} formalizes and analyzes key properties of tabular data principled in the relational data model, such as column-order insignificance. In this work, we focus on assessing the robustness on similar properties in the reasoning capabilities through the LLMs' generated responses. In this realm, \citet{singha2023tabular}, evaluate various LLMs on their tabular understanding capabilities under noisy tabular inputs and variations in formatting of tabular data in prompts. While their robustness assessments cover realistic characteristics such as permutations in column-order, we include more properties and assess more complex reasoning capabilities of LLMs.

\paragraph{Evaluation Metrics for Tabular Reasoning}
The TARGET benchmark~\cite{ji2024target} focuses on evaluating table retrieval methods in open-domain querying over structured data. They surface issues in the reliability of free-form text evaluation metrics such as SacreBleu and BERT-score, as ground-truth answers in tabular reasoning datasets are small text snippets or exact values while LLMs generate longer outputs which challenges such metrics. Other work~\cite{sui2024table} intends to remedy the evaluation problem by forcing an LLM to output a singular answer in a structured format. While relying on an LLMs' structured output generation and response parsing are prone to error, ground-truth answers are often sentences, albeit short, and not single values~\cite{chen2020open}. An alternative procedure, adopted in the TQA-Bench~\cite{qiu2024tqa}, is to include the ground-truth answer in multiple-choice options and let the LLM select an option. The validity hence reliability of this evaluation approach is questionable as it leaks the ground-truth answer in the input prompt.

\section{The TQA-Bench and Revisions}\label{sec:tqa-bench-revisions}

Here, we explain the tabular reasoning tasks included in the TQA-Bench that we use to assess the reliability of evaluation metrics as well as the robustness of the tabular reasoning capabilities of LLMs. We explain revisions we made to invalid queries, and tabular inputs to gain granular insights.

\subsection{TQA-Bench reasoning tasks}
The TQA-Bench~\cite{qiu2024tqa} provides a benchmark for tabular reasoning capabilities of three complexities: 1) lookup queries, 2) aggregation, and 3) complex calculations. Specifically, the TQA-Benchmark evaluates three different levels of reasoning complexities as follows:

\paragraph{Lookup queries} These queries involve simple entity extraction based on one or two direct conditions. For example, \textit{``What is the description of air carrier 20398?''} (Entity lookup) or \textit{``Which Horror movie gets the highest budget?''} (Top selection) require the model to retrieve a value from a column given a key or set of values from the same or another table. In multi-table settings, the challenge lies in resolving foreign key relationships.

\paragraph{Aggregation queries} These queries require calculations over filtered table segments. Examples include \textit{``How many airlines land in Flint, MI: Bishop International?''} (Count), \textit{``What is the total flight delay (DEP\_DELAY) from ORD?''} (Sum), and \textit{``What is the average arrival delay for flights landing at FNT?''} (Average). These tasks test the model's ability to perform basic numeric operations while managing filters and joins.

\paragraph{Complex calculations} These queries go beyond basic aggregation by requiring operations between multiple fields or statistical analysis. For instance, \textit{``What is the average total delay (ARR\_DELAY - DEP\_DELAY) for Envoy Air (MQ)?''} (Subtraction) and \textit{``What is the correlation between departure and arrival delays for flights with delays over -9 minutes?''} (Correlation) assess deeper reasoning by requiring chained arithmetic, statistical computation, and multi-step reasoning.

\subsection{TQA-Bench Revisions}

We leverage the tabular data and query generation methods from TQA-Bench but make two adjustments which we describe here: 1) we improve the validity of the queries, and 2) we downscale the tabular data inputs to yield more granular insights.

\paragraph{Query Refinements}
We updated some of the existing question templates, as they lead to unnatural questions such as ``\textit{Where is the 16S21E21G001S?}''. For cases such as this, we adapted the templates to be more precise and in line with natural questions. For instance, the question ``\textit{Where is the 16S21E21G001S?}'' is updated to ``\textit{In which county is the the station with the full name/id 16S21E21G001S?}``. These updates also ensured that there is only one logical answer which can be extracted from the available tables, as the original question could have also referred to the longitude and latitude columns of the respective dataset.

\paragraph{Tabular Data Downscaling}
While TQA-Bench evaluates multi-table reasoning capabilities with relatively large and multiple tables resulting in context sizes from 8K to 128K tokens. Our preliminary experiments revealed significant challenges in reasoning capabilities already with smaller table sizes, motivating the downsizing of context sizes to 1K, 2K, 4K, 6K and 8K to obtain more granular insights. To do so, we employ the scaling method introduced by TQA-Bench to truncate and segment tables while preserving their structural and relational integrity~\cite{qiu2024tqa}.

\section{Towards Reliable Evaluation of Tabular Reasoning Capabilities}

\begin{table*}[ht!]
\caption{Below query examples from the OTTQA benchmark~\cite{chen2020open} illustrate the difficulty of evaluating long-form LLM-generated answers against ground-truth answers (exact values or short text snippets) with existing text metrics such as the SacreBleu (SB) and BERT-score (BS) metrics, which fail to distinguish between incorrect versus correct answers. We find that an LLM-as-a-judge is a more reliable metric for evaluating tabular reasoning capabilities in these cases.}
\label{tab:metric-issues-examples}
\begin{tabularx}{\textwidth}{X c c c c}
    \toprule
     \multicolumn{5}{>{\setlength{\hsize}{2\hsize}\addtolength{\hsize}{2\tabcolsep}}X}{\textbf{Question:} \textit{According to the 2016 census, what is the population of the city that is home to the Territorial Administration Building designed by Thomas W. Fuller?}} \\
    \midrule
    \textbf{Generated answer} &\textbf{Answer} & \textbf{SB score} & \textbf{BS score} & \textbf{LLM} \\
     \midrule
      The Territorial Administration Building is located in Dawson City, Yukon. According to the 2016 Canadian census, the population of Dawson City was \colorbox{YellowGreen}{1,375}. & 1,375 &  1.4 & 0.83 & \colorbox{YellowGreen}{correct}\\
     According to the 2016 census, the population of the city that is home to the Territorial Administration Building designed by Thomas W. Fuller is \colorbox{Salmon}{45,000}. & 1,375 &  0.0 & 0.81 & \colorbox{Salmon}{incorrect}\\
     \bottomrule
\end{tabularx}
\end{table*}
\begin{table*}[ht!]
\begin{tabularx}{\textwidth}{X c c c c}
    \toprule
     \multicolumn{5}{>{\setlength{\hsize}{2\hsize}\addtolength{\hsize}{2\tabcolsep}}X}{\textbf{Question:} \textit{How long has the team that qualified for the 1999 FIFA Confederations Cup in June existed?}}  \\
     \midrule
     \textbf{Generated answer} & \textbf{Answer} & \textbf{SB score} & \textbf{BS score} & \textbf{LLM} \\
    \midrule
      The team that qualified for the 1999 FIFA Confederations Cup in June was Germany, who secured their qualification on 30 June 1996. To determine how long they had existed as a team by June 1999: The Germany national football team was officially \colorbox{YellowGreen}{founded in 1908}. From 1908 to 1999 is 91 years. Answer: The Germany national team had existed for 91 years by June 1999. & since 1908 & 0.5 & 0.81 & \colorbox{YellowGreen}{correct} \\
     The team that qualified for the 1999 FIFA Confederations Cup in June was Germany. Germany has existed as a national team \colorbox{Salmon}{since 1974}, so by June 1999, they had existed for 25 years. & since 1908 & 1.0 & 0.82 & \colorbox{Salmon}{incorrect}\\
     \bottomrule
\end{tabularx}
\vspace{-0.5cm}
\end{table*}

Evaluating multi-table reasoning, and generally free-form question answering, still is an open challenge~\cite{ji2024target, xu2023critical}. While context and reasoning traces of LLM-generated answers are useful, they complicate evaluation when ground-truth answers are short and exact, as is the case in typical tabular tasks. Table~\ref{tab:metric-issues-examples} illustrates this issue for two example queries from the popular OTTQA dataset for table question answering~\cite{chen2020open} along with their ground-truth answers. When the LLM-generated answers are evaluated against the short ground-truth answers by two free-form text metrics, SacreBleu~\cite{post-2018-call} and BERT-score~\cite{Zhang2020BERTScore:}, their scores are inconclusive. For example, for the queries in Table~\ref{tab:metric-issues-examples}, the SB score for a correct generated answer is higher for the first query (1.4) but lower than the correct generated answer for the second query (0.5) for which the incorrect generated answer is closer to the correct answer for the first query (1.0). The BERT-score (BS) reflects mainly textual similarity, and shows barely any differences for different numeric values included in the response: its value is 0.81 for an incorrect as well as a correct generated answer. In what follows, we study the reliability of these different metrics in detail.

\paragraph{LLM-generated answers}
% ADD: table .
We adopt a synthetic procedure to maximize the likelihood of (in)correctness of the LLM-generated answers, which we refer to as \textit{generated answers}. In total, we extract 350 questions from the question database in the TQA-Bench tabular reasoning benchmark. We only provide a single row as context to the LLM to generate its answer, while forcing it to only use the table data and not its memory. For \textit{correct} generated answers, we provide the row that contains the ground-truth cell value for the lookup task as context to generate the response. For \textit{incorrect} answers, we also provide the row that contains the ground-truth cell, but replace the ground-truth cell value with a random but different value from the same column, resulting in a factually incorrect but still realistic generated answer. Following this procedure, we extract 175 correct and 175 incorrect generated answers.

To validate the LLM-generated answers, we check the correctness of the 350 generated reference answers against the ground-truth answers (original cell values) through human annotation. The human evaluation reveals that approximately 93.75\% of the generated correct answers are indeed accurate, while 98.26\% of the generated incorrect answers are indeed incorrect. These results confirm the reliability of our procedure for creating the LLM-generated answer dataset.

\begin{figure}[h!]
  \centering
  \begin{subfigure}{\linewidth}
    \centering
    \includegraphics[width=0.7\linewidth]{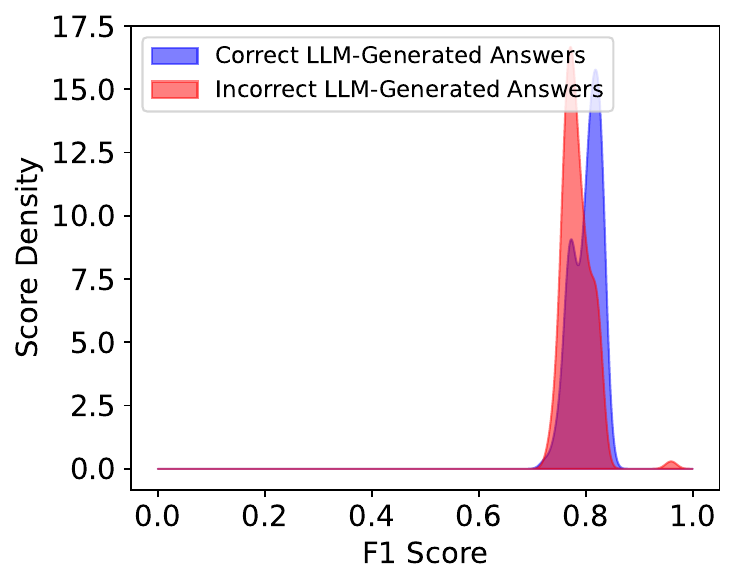}
    \caption{BERT Score distribution using Kernel Density Estimation (KDE) of the incorrect and correct generated answers. Scores close to 1 indicate stronger semantic similarity between LLM-generated and ground-truth answers.}
    \label{fig:bert_scores}
  \end{subfigure}
  \hfill
  \begin{subfigure}{\linewidth}
    \centering
    \includegraphics[width=0.7\linewidth]{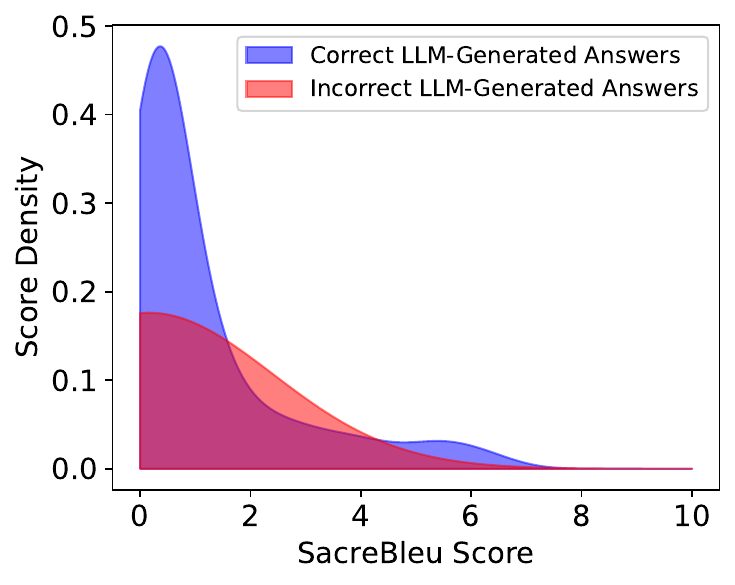}
    \caption{SacreBleu Score distribution using Kernel Density Estimation (KDE). Higher scores indicate better n-gram overlap between LLM-generated and ground-truth answers.}
    \label{fig:bleu_scores}
  \end{subfigure}
  \caption{Distribution of SacreBleu and BERT-scores obtained by comparing LLM-generated answers, from which we know their (in)correctness, against ground-truth answers. To be reliable as a metric, the distributions should be clearly separable, which is not the case for both metrics.}
  \label{fig:metrics}
\end{figure}

\paragraph{Free-form text evaluation metrics}
\label{sec:free-form-text-metrics}
Using the LLM-generated answers and ground-truth answers, we inspect the reliability of two commonly used free-form text metrics: SacreBleu~\cite{post-2018-call} and BERT-score~\cite{Zhang2020BERTScore:}. SacreBLEU is a standardized version of the BLEU score that measures n-gram overlap between generated and reference texts, while the BERT-score leverages BERT embeddings to compute similarity based on token-level semantic matching. Our analysis reveals that neither the BERT-score nor SacreBleu metric provide a reliable signal for evaluating the correctness of generated answers. To visualize the reliability of these scores, we used Kernel Density Estimation (KDE) to estimate their distributions. For BERT-score, due to tight clustering of values, the KDE can exceed 1, while the more dispersed SacreBLEU values result in lower KDE peaks.
The distributions of scores for correct and incorrect LLM-generated answers, when compared with ground-truth answers, exhibit significant overlap making them indistinguishable (Figures~\ref{fig:bert_scores} and~\ref{fig:bleu_scores}). The inseparability between these distributions illustrates the unsuitability of these metrics for evaluating the accuracy of long-form answers against concise ground-truth answers.

\paragraph{LLM-as-a-judge} Recently, LLMs have emerged as a useful evaluation metric for free-form text, termed as LLM-as-a-judge evaluation~\cite{zheng2023judging}. This approach to be particularly well-suited for tabular reasoning evaluation, where generated answers are often long- and free-form text compared to, for example OTTQA, where answers are short text snippets or single (numeric) values. We propose using an LLM-as-a-judge for evaluating tabular reasoning through LLMs, as this allows us to keep the generation close to real-world usage, where users expect models to generate complete answers rather than forcing a single-valued answer (which does not always correspond to the ground-truth answer) or choose from predefined options as in TQA-Bench~\cite{qiu2024tqa}. Second, relying on forced answer formats --accommodating multiple-choice or string-matching-- can be brittle, especially for smaller models that may produce slightly misformatted outputs or fail to follow constraint templates \cite{liu2024llms}.

To understand the reliability of the LLM-as-a-judge for evaluating tabular reasoning, we first evaluate the performance of the LLM-as-a-judge. Specifically, we devise reference-guided grading~\cite{zheng2023judging} and let the LLM compare between the LLM-generated answer against the ground-truth answer. Specifically, we use Qwen2.5 (32B parameters) and assess if its generated answer matches the ground-truth answer based on a structured prompt, and outputs \texttt{yes} or \texttt{no}, as follows:

\begin{table}[h]
\begin{prompt}
\footnotesize{
    When it comes to the following question: \\
    
    \textbf{Question:} {\color{blue}\{Question\}}\\

    does the answer "{\color{blue}\{Answer\}}" match the expected response value of the correct answer "{\color{blue}\{Correct Value\}}"?\\
    
    Consider that if the answer is None, it means that the value could not be found in the table. Please conclude your answer with 'answer correct: yes/no'
}
\end{prompt}
\end{table}
\vspace{-0.5cm}

\begin{table}[h]
\centering
\caption{Evaluation of the LLM-as-a-judge procedure on the human-annotated dataset. While the LLM slightly underestimates correctness -- 4.2\% of correct answers are judged to be incorrect -- we observe a strong alignment between predicted and actual (in)correctness with an accuracy more than 95\%.}
\label{tab:confusion_matrix_llm}
\renewcommand{\arraystretch}{1.2}
\resizebox{\columnwidth}{!}{%
    \begin{tabular}{lcc}
    \toprule
     & \textbf{Pred. Incorrect} & \textbf{Pred. Correct} \\ \hline
    \textbf{Actual Incorrect} & 99.2\% & 0.8\% \\ 
    \textbf{Actual Correct} & 4.2\% & 95.8\% \\ 
    \bottomrule
    \end{tabular}
}
\end{table}

The results of our evaluation (Table~\ref{tab:confusion_matrix_llm}) demonstrates that the LLM-as-a-judge yields a high accuracy, identifying 95.8\% of correct answers as correct, and 98\% of the incorrect answers as such. Notably, the absence of false positives (0.8\%) highlights the model's reliability in avoiding incorrect classifications of negative cases as positive.

\section{On the Realistic Tabular Reasoning Capabilities of LLMs}
% start with reflecting the scaling motivation by two plots showing performance decrease -> add a large square for some consistent model on 8K to indicate gap with tqa-bench.
% duplicate values: table with models on rows, tasks on columns with performance results split into baseline | duplicate "subcolumns".
% missing values: models on rows with tasks average/sum/lookup on "subrows", with multi-level columns "danger zone": undesired behavior, incorrect. and "green zone": desired behavior, correct answer -> ideally at least for qwen and gpt-4o(-mini).
% structural variations: do include some numbers, but we can conclude while structural variation does not affect performance here, it might in future models hence we can test for this. moreover, point to target benchmark for influence on position of answer in the context -- this deeper analysis is for future work.
% interpretation:
% - nuancing refraining from answering versus incorrectness versus table size.
% - differences across models?
% - error analysis.

Here, we first examine how well LLMs can perform tabular reasoning tasks using the downscaled TQA-Bench data (as discussed in Section~\ref{sec:tqa-bench-revisions}) and the more reliable LLM-as-a-judge evaluation, and highlight new insights contextualized in the TQA-Bench. Then, we extend the benchmark by formalizing tabular characteristics commonly found in practice, such as missing values, and measure the robustness of LLMs for such realistic variations.

\subsection{LLM Selection and Prompts}
% section naming is still work in progress

\paragraph{LLMs for analysis and evaluation}
We conduct our analysis on a diverse set of models to ensure comprehensive evaluation. We include publicly available models Qwen2.5~\cite{yang2024qwen2}, Llama3.1~\cite{grattafiori2024llama}, DeepSeek-R1~\cite{guo2025deepseek}, and Mistral~\cite{jiang2023mistral7b}. In all cases, we select the 7B parameter versions (except llama3.1 with 8B) of the models and utilize the same prompt structure. We also include the proprietary GPT-4o-mini model (version 2024-07-18)~\cite{hurst2024gpt} representing a state-of-the-art larger general-purpose model.

For evaluation with the LLM-as-a-judge procedure we, again, devise Qwen2.5 (32B parameters) for its strength in generating structured outputs, and use the same prompt introduced in section \ref{sec:free-form-text-metrics}.

\paragraph{Tabular reasoning prompt} For evaluating tabular reasoning capabilities of the LLMs, we adopt a structured prompt template inspired by \citet{qiu2024tqa} to guide question answering based on tabular data\footnote{We inspected accuracy variance across templates and didn't observe a significant difference.}. The prompt instructs the LLM to use the information from the provided single or multiple tables to answer a given question. Each table is presented with a title and its contents. The structure of the prompt template is as follows:
\vspace{-0.2cm}

\begin{prompt}
\vspace{-0.1cm}
\footnotesize{
    Answer the question based on these tables:\\
    
    \textbf{Table:} {\color{blue}\{Table 1\}}
    
    \textbf{Table:} {\color{blue}\{Table 2\}}\\
    
    \textbf{Question:} {\color{blue}\{Question\}}\\
    
    This question has only one correct answer. Please break down the question, evaluate each option, and explain why it is correct or incorrect. Conclude with your final answer.
}
\vspace{-0.1cm}
\end{prompt}

\subsection{Insights on Down-scaled Tabular Inputs}

\begin{table*}[ht!]
\centering
\caption{Accuracies of LLMs across reasoning tasks for tables with token size 4k. Generally, LLMs can reasonably do basic entity lookups, but show large deficits in more complex reasoning tasks such as calculating averages and correlations. As expected given the larger model size, GPT-4o-mini shows best performance across tasks.}
\label{tab:model-performance}
% \renewcommand{\arraystretch}{1}
% \resizebox{\textwidth}{!}{%
\small{
\begin{tabular}{lccccccc}
   \textbf{Model}  & \textbf{Entity lookup} & \textbf{Top selection} & \textbf{Average} & \textbf{Count} & \textbf{Subtraction} & \textbf{Sum} & \textbf{Correlation} \\
    \midrule
    Llama3.1 & 49.75 & 22.96 & 16.33 & 28.79 & 20.92 & 16.08 & \colorbox{Salmon}{4.90} \\
    Mistral & 28.14 & \colorbox{Salmon}{14.00} & 9.14 & \colorbox{Salmon}{20.20} & 5.58 & 8.50 & 12.75 \\
    Qwen2.5 & 29.00 & 15.00 & 11.28 & 36.92 & 7.22 & 12.50 & 8.65 \\
    Deepseek-r1 & \colorbox{Salmon}{20.00} & 14.56 & \colorbox{Salmon}{7.64} & 25.95 & \colorbox{Salmon}{3.90} & \colorbox{Salmon}{8.18} & 20.48 \\
    GPT-4o-mini & \colorbox{YellowGreen}{68.72} & \colorbox{YellowGreen}{44.62} & \colorbox{YellowGreen}{32.83} & \colorbox{YellowGreen}{49.75} & \colorbox{YellowGreen}{36.73} & \colorbox{YellowGreen}{35.86} & \colorbox{YellowGreen}{24.04} \\
    \bottomrule
\end{tabular}
}
% \begin{tabular}{lccccccc}
%   \toprule
%   \textbf{Model}  & \textbf{Average} & \textbf{Correlation} & \textbf{Count} & \textbf{Subtraction} & \textbf{Top selection} & \textbf{Entity lookup} & \textbf{Sum} \\
%   \midrule
%   \textbf{llama3.1} & 16.33 & \colorbox{Salmon}{4.90} & 28.79 & 20.92 & 22.96 & 49.75 & 16.08 \\
%   \textbf{qwen2.5} & 11.28 & 8.65 & 36.92 & 7.22 & 15.00 & 29.00 & 12.50 \\
%   \textbf{mistral} & 9.14 & 12.75 & \colorbox{Salmon}{20.20} & 5.58 & \colorbox{Salmon}{14.00} & 28.14 & 8.50 \\
%   \textbf{deepseek-r1} & \colorbox{Salmon}{7.64} & 20.48 & 25.95 & \colorbox{Salmon}{3.90} & 14.56 & \colorbox{Salmon}{20.00} & \colorbox{Salmon}{8.18} \\
%   \textbf{gpt-4o-mini} & \colorbox{YellowGreen}{32.83} & \colorbox{YellowGreen}{24.04} & \colorbox{YellowGreen}{49.75} & \colorbox{YellowGreen}{36.73} & \colorbox{YellowGreen}{44.62} & \colorbox{YellowGreen}{68.72} & \colorbox{YellowGreen}{35.86} \\
%   \bottomrule
% \end{tabular}
% }
\vspace{-0.3cm}
\end{table*}

% \begin{table*}[ht!]
% \centering
% \renewcommand{\arraystretch}{1}
% \resizebox{\textwidth}{!}{%
% \begin{tabular}{lccccccc}
%   \toprule
%   \textbf{Model}  & \textbf{Average} & \textbf{Correlation} & \textbf{Count} & \textbf{Subtraction} & \textbf{Top selection} & \textbf{Entity lookup} & \textbf{Sum} \\
%   \midrule
%   \textbf{llama3.1} & 16.33 & \colorbox{Salmon}{4.90} & 28.79 & 20.92 & 22.96 & 49.75 & 16.08 \\
%   \textbf{qwen2.5} & 11.28 & 8.65 & 36.92 & 7.22 & 15.00 & 29.00 & 12.50 \\
%   \textbf{mistral} & 9.14 & 12.75 & \colorbox{Salmon}{20.20} & 5.58 & \colorbox{Salmon}{14.00} & 28.14 & 8.50 \\
%   \textbf{deepseek-r1} & \colorbox{Salmon}{7.64} & 20.48 & 25.95 & \colorbox{Salmon}{3.90} & 14.56 & \colorbox{Salmon}{20.00} & \colorbox{Salmon}{8.18} \\
%   \textbf{gpt-4o-mini} & \colorbox{YellowGreen}{32.83} & \colorbox{YellowGreen}{24.04} & \colorbox{YellowGreen}{49.75} & \colorbox{YellowGreen}{36.73} & \colorbox{YellowGreen}{44.62} & \colorbox{YellowGreen}{68.72} & \colorbox{YellowGreen}{35.86} \\
%   \bottomrule
% \end{tabular}
% }
% \caption{Performance of various models across different reasoning tasks for the tables with token size 4k.}
% \label{tab:model-performance}
% \vspace{-0.3cm}
% \end{table*}

\begin{figure}[ht!]
  \centering
  \includegraphics[width=0.68\linewidth]{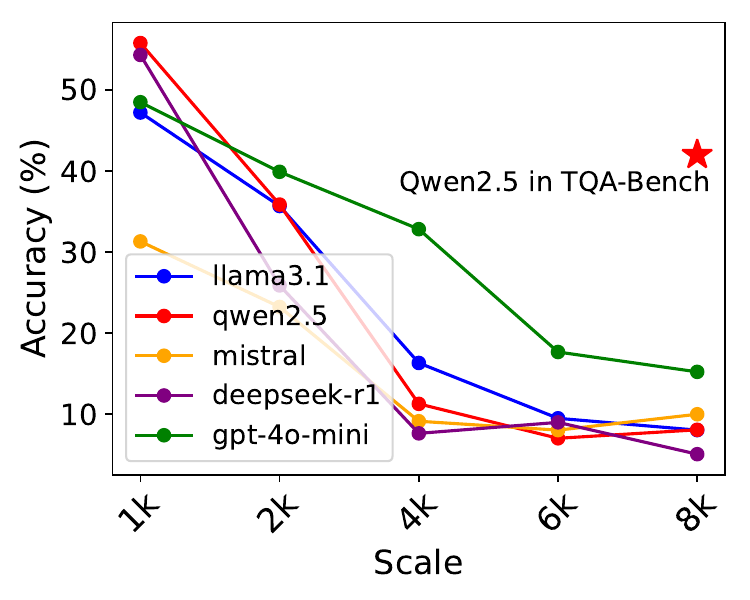}
  \caption{Performance of LLMs on calculating the \textit{average} of columns, across varying table sizes.  The accuracy of all models gradually decreases with table size.}
  \label{fig:average}
  \vspace{-0.5cm}
\end{figure}

\begin{figure}[ht!]
  \centering
  \includegraphics[width=0.68\linewidth]{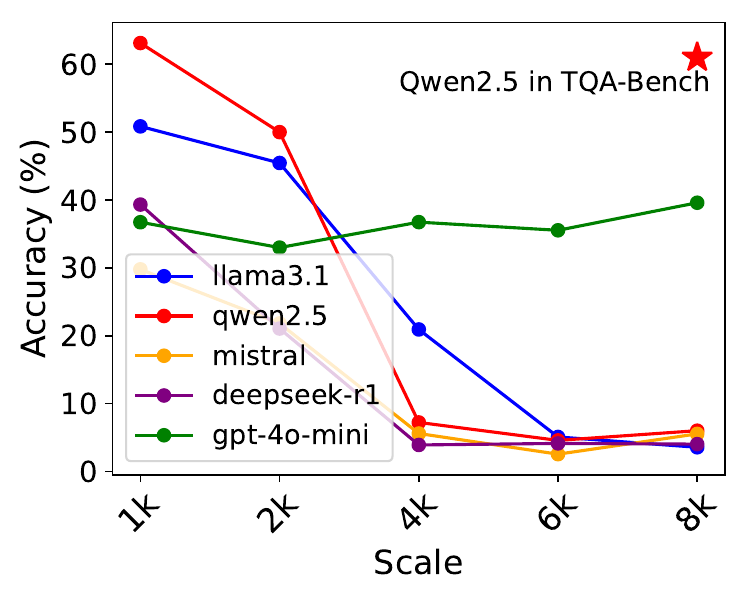}
  \caption{Performance of LLMs on calculating a \textit{subtraction} across columns, across varying table sizes. The accuracy significantly drops after 4K input size except for GPT-4o-mini.}
  \label{fig:difference}
  \vspace{-0.4cm}
\end{figure}

\paragraph{Accuracy over various table sizes} Our analysis of the TQA-Bench questions and down-scaled tabular inputs shows that the performance of the LLMs decreases as the tabular input increases in size. This is particularly evident in the \textit{average} and \textit{subtraction} tasks (Figure \ref{fig:average} and \ref{fig:difference}). The only exception is GPT-4o-mini, which achieves a steady performance across table sizes for most tasks, and is generally the best model for tabular reasoning tasks. Furthermore, our results indicate that LLMs struggle particularly with more complex reasoning tasks, such as calculating correlation and subtraction, where the performance is significantly lower compared to the simpler tasks like counting and lookups. A comprehensive overview of the accuracy performances across all models and tasks can be found in appendix \ref{sec:full-benchmark}.

\vspace{-0.15cm}
\paragraph{Realistic LLM performance on TQA-Bench}
We also plot the accuracy of the Qwen2.5 model (7B params), as found by the TQA-Bench multiple-choice evaluation, for the \textit{average} and \textit{subtraction} tasks for the 8K sized tabular inputs (\textcolor{red}{\ding{72}} in Figures \ref{fig:average} and \ref{fig:difference}). Our LLM-as-a-judge evaluation of open-form answers unveils a significant difference in accuracy of 30\% and 60\% for \textit{average} and \textit{subtraction} calculations, respectively, compared to multiple-choice answering. 
This insight underscores the importance of evaluating models in open-ended form to better understand their true reasoning abilities.
Across models, we generally observe relatively stronger capabilities in entity lookups, while selecting a range (\textit{top selection}) is more challenging (Table~\ref{tab:model-performance}). LLMs show larger deficits in more complex aggregation tasks, such as calculating averages, while the relatively basic task of subtraction generally appears most challenging.

\vspace{-0.15cm}
\paragraph{Model-specific incompatible behaviors}
During analysis, we observed notable behaviors in how models approached tabular reasoning tasks. For instance, DeepSeek-R1 often struggles with coherence in its chain of thought outputting extracts like ``\textit{Wait no—the data doesn't show that. Wait I'm getting confused.}'', leading to incomplete or inconsistent reasoning.  Llama3.1, on the other hand, occasionally fails to generate any meaningful output, and effectively ``breaks'' under certain conditions, particularly on larger tables or complex queries. Additionally, both Qwen2.5 and Llama3.1 attempt to generate Python code snippets to compute answers, rather than directly providing the response.

\subsection{Real-world characteristics of tabular data}
We analyze how robust LLMs can reason over tabular data through their generation capabilities, in the presence of three variations in the tabular inputs: missing values, duplicate entities, and column permutations. These variations are common in practice and resemble either data quality issues or valid permutations. In what follows, we elaborate on the desired behavior which goes beyond accuracy~\cite{xu2023critical}, for each characteristic. In order to instill these characteristics in the tabular inputs, we adapt the symbolic extension of TQA-Bench to generate new tasks. Symbolic extension works by manually creating prompt templates and functions for calculating the ground-truth answers, and using them to automatically generate combinations of (question, ground-truth)-pairs.

\vspace{-0.15cm}
\paragraph{Missing Values} Due to incomplete data collection or errors during data entry, tables often contain missing values, leading to incomplete information~\cite{little2019statistical}. This has been a longstanding issue for predictive ML, as missing values can distort analysis and lead to unreliable results \cite{emmanuel2021survey}. To reflect this in the data, we first identify the cells needed to generate the answers and randomly remove one of the relevant cells. We recalculate the ground-truth answer by, effectively, setting the missing value to 0. The LLM-as-a-judge evaluates two behaviors: the model’s ability to produce the correct answer despite the missing information (\textbf{Accuracy}), and its capacity to explicitly acknowledge the absence of relevant data (\textbf{Acknowledgement}). These criteria reflect how well the model navigates incomplete data—whether it can reason effectively with what’s available—and transparently communicate the limitations introduced by missing values.

\vspace{-0.15cm}
\paragraph{Duplicate Entities} While duplicate entities are in violation with the relational data model~\cite{codd1979extending}, we often find duplicate rows in tables. We simulate this by randomly selecting rows and replicating them at random points in the same table. These entities are intended to be ignored. The LLM-as-a-judge then evaluates two desired behaviors: whether the correct answer has been generated despite duplicates values (\textbf{Accuracy}), hence ignoring duplicate values, and whether the model explicitly acknowledges the duplicates in its response (\textbf{Acknowledgement}).

\vspace{-0.15cm}
\paragraph{Structural Variation}
While tables within some contexts might reflect a typical column or row order, hence bias an LLM through its training data, the structural order of tables is insignificant in the relational data model~\cite{codd1979extending}. In line with prior work for examining robustness of table embeddings~\cite{cong2023observatory}, we extract different permutations of the same tables by shuffling their rows and columns. The desired behavior is that the answer is not affected by different column order permutations of the input tables. Therefore, we evaluate if the answer remains unchanged.

\subsection{Robustness to Realistic Variation in Tables}

%Here, we present the insights on the robustness of the LLMs for realistic variations in tabular inputs.

\paragraph{Missing Values} We observe some mixed behaviors in the presence of missing values across tasks (Table~\ref{tab:results-missing-values}). For summation, we observe a significant drop in \textbf{accuracy} when missing values are present for llama3.1, which achieves only 8\% accuracy compared to 16\% in the baseline. In contrast, qwen2.5 actually shows improvements, particularly for entity lookups where it shows an accuracy of 43\% compared to 29\%.
%In general, there is no significant and consistent pattern in terms of accuracy.
%,as the performance differences could also be explained by randomness in the data.
This may be due to the model's (desirable) behavior as it refuses to answer if the value to-be-looked-up is missing, which is treated by the judge as a correct response.
% Overall, while we do not observe a clear and consistent pattern in accuracy across tasks, small variations—on the order of a few percentage points—can be attributed to randomness, given that each task and model configuration includes a few hundred prompts.

At the same time, we find that the models often \textbf{acknowledge} missing values in their responses, with both models achieving around 44\% in the sum task and 51\% in the average task. This suggests that while models may struggle with accuracy, they are still able to recognize and communicate the presence of missing values in their answers. Still, even on this metric, the models do not behave reliably enough for most practical use cases.

\begin{table}[h!]
\vspace{-0.15cm}
\caption{Results of the Missing Value perturbation across three aggregation tasks (average, sum and entity lookups) with table size 4k.}
\label{tab:results-missing-values}
\vspace{-0.1cm}
\renewcommand{\arraystretch}{1}
\resizebox{\columnwidth}{!}{%
    \begin{tabular}{ll ccc}
    \toprule
    \textbf{Task} & \textbf{Model} & \textbf{Baseline} & \textbf{Acc.} & \textbf{Acknow.} \\
    \hline
    \textbf{Entity lookup} & \textbf{llama3.1} & 50\% & 47\% & 57\% \\
     & \textbf{qwen2.5} & 29\% & 43\% & 60\% \\
    \textbf{Sum} & \textbf{llama3.1} & 16\% & 8\% & 44\% \\
     & \textbf{qwen2.5} & 13\% & 16\% & 44\% \\
    \textbf{Average} & \textbf{llama3.1} & 17\% & 11\% & 51\% \\
     & \textbf{qwen2.5} & 11\% & 19\% & 51\% \\
    \bottomrule
    \end{tabular}
}
\vspace{-0.3cm}
\end{table}

\paragraph{Duplicate Entities} When it comes to dealing with duplicate entities, the trends displayed in table \ref{tab:results-duplicates} overall are quite similar to dealing with missing values, showing a significant decrease in \textbf{accuracy} in most tasks. A notable observation is that models are less likely to \textbf{acknowledge} duplicate values, compared to missing values.

\begin{table}[h!]
\centering
\vspace{-0.15cm}
\caption{Results of the Duplication perturbation across two advanced tasks (average, sum) at table size 2k. LLMs typically struggle with duplicate values, and fail to acknowledge duplication in their response.}
\label{tab:results-duplicates}
\vspace{-0.1cm}
\renewcommand{\arraystretch}{1}
\resizebox{\columnwidth}{!}{%
    \begin{tabular}{ll ccc}
    \toprule
    \textbf{Task} & \textbf{Model} & \textbf{Baseline} & \textbf{Acc.} & \textbf{Acknow.} \\
    \hline
    \textbf{Sum} & \textbf{llama3.1} & 41\% & 20\% & 27\% \\
     & \textbf{qwen2.5} & 30\% & 31\% & 8\% \\
    \textbf{Average} & \textbf{llama3.1} & 36\% & 17\% & 11\% \\
     & \textbf{qwen2.5} & 36\% & 20\% & 6\% \\
    \bottomrule
    \end{tabular}
}
\vspace{-0.33cm}
\end{table}

\paragraph{Structural Variations} We find that structural variations, such as column shuffling, have only a small impact on model performance in most cases (Table \ref{tab:results-col-shuffling}), in contrast to the significant performance decline observed with missing values or duplicate entities. Interestingly, the robustness to column order varies across models and tasks—some exhibit resilience, while others are mildly affected. This shows a divergence from previous findings in embedding-based studies \cite{cong2023observatory}, which reported a sensitivity to column order in the representation space.

\begin{table}[h!]
\centering
\vspace{-0.15cm}
\small{
\caption{Impact of column shuffling on select reasoning tasks with a table size of 2k, showing difficulties for aggregation queries particularly for the llama model.}
\label{tab:results-col-shuffling}
\vspace{-0.1cm}
    \begin{tabular}{ll ccc}
    \toprule
    \textbf{Task} & \textbf{Model} & \textbf{Baseline} & \textbf{Acc.} \\
    \hline
    \textbf{Entity lookup} & \textbf{llama3.1} & 50\% & 46\%\\
      & \textbf{qwen2.5} & 42\% & 34\% \\
    \textbf{Sum} & \textbf{llama3.1} & 41\% & 30\% \\
      & \textbf{qwen2.5} & 30\% & 28\% \\
    \textbf{Average} & \textbf{llama3.1} & 36\% & 28\% \\

    & \textbf{qwen2.5} & 36\% & 32\% \\
    \bottomrule
    \end{tabular}
    }
    \vspace{-0.3cm}
\end{table}

\section{Conclusion}
\vspace{-0.15cm}
While recent studies suggest LLMs exhibit reasonable tabular reasoning abilities beyond natural language tasks, these studies often lack reliable evaluations and robustness checks, prompting our study into how well LLMs \textit{truly} reason over tabular inputs.
First, we surface limitations of common free-form text evaluation metrics, such as SacreBleu, which fail to distinguish between correct and incorrect answers in tabular reasoning tasks. We demonstrate that an LLM-as-a-judge is more reliable for this purpose. A revised evaluation of an existing benchmark with the LLM-as-a-judge unveils a significant deficit in tabular reasoning capabilities of LLMs.
Second, we analyze the robustness of tabular reasoning capabilities of LLMs through queries of various complexities and find that they can be sensitive to realistic variations like missing values, even for relatively simple tasks. Moreover, we find that LLMs insufficiently acknowledge such undesired variations risking errors in downstream interpretation. These findings underscore the need for further advancements in LLM architectures and training to improve their robustness to real-world tabular data.

\section*{Acknowledgments} \vspace{-0.2cm}This work was partially funded by grants from NWO (NGF.1607.22.045) and SAP.
\vspace{-0.35cm}
\bibliography{custom}

\appendix
\onecolumn
\label{sec:appendix}

\section{Full Benchmark}
\label{sec:full-benchmark}

This table provides a comprehensive overview of the performance of various LLMs across different reasoning tasks (e.g., entity lookup, top selection, average, etc.) for tabular data of varying sizes (1k to 8k tokens).

\vspace{0.5cm}

\renewcommand{\arraystretch}{1}
\resizebox{.97\columnwidth}{!}{%
\begin{tabular}{lrrrrrrrr}
\toprule
\textbf{Size} & \textbf{Model} & \textbf{Entity lookup} & \textbf{Top Selection} & \textbf{Average} & \textbf{Count} & \textbf{Subtraction} & \textbf{Sum} & \textbf{Correlation} \\
\hline
\multicolumn{9}{l}{\textbf{1k}} \\
 & llama3.1 & 43.43 & 30.15 & 47.21 & 59.30 & 50.85 & 52.02 & 18.64 \\
 & mistral & 30.15 & 31.31 & 31.31 & 47.50 & 29.78 & 33.67 & 36.75 \\
 & qwen2.5 & 49.24 & 30.30 & 55.78 & 65.15 & 63.13 & 56.78 & 37.82 \\
 & deepseek-r1 & 34.01 & 29.80 & 54.31 & 69.19 & 39.33 & 50.00 & 47.46 \\
 & gpt-4o-mini & 41.62 & 36.18 & 48.48 & 70.56 & 36.72 & 54.50 & 45.38 \\
\multicolumn{9}{l}{\textbf{2k}} \\
 & llama3.1 & 76.38 & 49.49 & 35.68 & 44.95 & 45.45 & 41.21 & 11.21 \\
 & mistral & 58.08 & 36.18 & 23.23 & 28.00 & 21.81 & 17.17 & 16.82 \\
 & qwen2.5 & 66.33 & 42.86 & 35.86 & 53.27 & 50.00 & 30.26 & 22.64 \\
 & deepseek-r1 & 42.93 & 28.28 & 25.89 & 49.75 & 21.05 & 26.26 & 24.07 \\
 & gpt-4o-mini & 69.54 & 47.24 & 39.90 & 57.07 & 32.98 & 37.76 & 28.44 \\
\multicolumn{9}{l}{\textbf{4k}} \\
 & llama3.1 & 49.75 & 22.96 & 16.33 & 28.79 & 20.92 & 16.08 & 4.90 \\
 & mistral & 28.14 & 14.00 & 9.14 & 20.20 & 5.58 & 8.50 & 12.75 \\
 & qwen2.5 & 29.00 & 15.00 & 11.28 & 36.92 & 7.22 & 12.50 & 8.65 \\
 & deepseek-r1 & 20.00 & 14.56 & 7.64 & 25.95 & 3.90 & 8.18 & 20.48 \\
 & gpt-4o-mini & 68.72 & 44.62 & 32.83 & 49.75 & 36.73 & 35.86 & 24.04 \\
\multicolumn{9}{l}{\textbf{6k}} \\
 & llama3.1 & 21.11 & 15.15 & 9.50 & 19.60 & 5.08 & 7.50 & 4.00 \\
 & mistral & 11.56 & 12.00 & 8.04 & 16.00 & 2.54 & 4.06 & 8.82 \\
 & qwen2.5 & 16.16 & 8.00 & 7.04 & 19.50 & 4.57 & 6.53 & 9.80 \\
 & deepseek-r1 & 7.00 & 11.50 & 9.00 & 16.67 & 4.12 & 5.50 & 9.90 \\
 & gpt-4o-mini & 68.53 & 42.13 & 17.68 & 38.50 & 35.53 & 16.08 & 16.16 \\
\multicolumn{9}{l}{\textbf{8k}} \\
 & llama3.1 & 12.50 & 10.55 & 8.04 & 18.00 & 3.55 & 2.54 & 3.03 \\
 & mistral & 9.00 & 10.10 & 10.00 & 8.00 & 5.53 & 2.51 & 10.31 \\
 & qwen2.5 & 8.04 & 7.54 & 8.08 & 11.50 & 6.00 & 1.51 & 9.00 \\
 & deepseek-r1 & 8.04 & 12.00 & 5.08 & 13.00 & 4.02 & 4.00 & 9.09 \\
 & gpt-4o-mini & 64.00 & 35.86 & 15.23 & 32.16 & 39.59 & 14.80 & 11.22 \\
\bottomrule
\end{tabular}
}

\end{document}